\documentclass[10pt,twocolumn,letterpaper]{article}

\usepackage{cvpr}              

\usepackage{multirow}

\usepackage{amssymb}
\usepackage{pifont}
\newcommand{\cmark}{\ding{51}}%
\newcommand{\xmark}{\ding{55}}%

\usepackage{graphicx}
\usepackage{booktabs}

\usepackage{subcaption}

\usepackage{algorithm}
\usepackage{algpseudocode}

\definecolor{cvprblue}{rgb}{0.21,0.49,0.74}
\usepackage[pagebackref,breaklinks,colorlinks,allcolors=cvprblue]{hyperref}

\def\paperID{*****} 
\def\confName{CVPR}
\def\confYear{2025}

\title{Chitrakshara: A Large Multilingual Multimodal Dataset for Indian languages}

\author{Shaharukh Khan$^*$, Ali Faraz$^*$, Abhinav Ravi$^\dagger$, Mohd Nauman, Mohd Sarfraz, \\ Akshat Patidar, Raja Kolla, Chandra Khatri$^\dagger$, Shubham Agarwal$^\dagger$ \\ 
Krutrim AI, Bangalore, India\\
}

\begin{document}
\maketitle
\def\thefootnote{*}\footnotetext{Equal contribution}\def\thefootnote{\arabic{footnote}}
\def\thefootnote{$\dagger$}\footnotetext{Senior contributors. Contact: \{shaharukh.khan, shubham.agarwal1, abhinav.ravi\}@olakrutrim.com}\def\thefootnote{\arabic{footnote}}
\setcounter{footnote}{0}

\begin{abstract}


Multimodal research has predominantly focused on single-image reasoning, with limited exploration of multi-image scenarios. Recent models have sought to enhance multi-image understanding through large-scale pretraining on interleaved image-text datasets. However, most Vision-Language Models (VLMs) are trained primarily on English datasets, leading to inadequate representation of Indian languages. To address this gap, we introduce the Chitrakshara dataset series, covering 11 Indian languages sourced from Common Crawl. It comprises (1) Chitrakshara-IL, a large-scale interleaved pretraining dataset with 193M images, 30B text tokens, and 50M multilingual documents, and (2) Chitrakshara-Cap, which includes 44M image-text pairs with 733M tokens. This paper details the data collection pipeline, including curation, filtering, and processing methodologies. Additionally, we present a comprehensive quality and diversity analysis to assess the dataset’s representativeness across Indic languages and its potential for developing more culturally inclusive VLMs.

\end{abstract}

\section{Introduction}
\label{sec:intro}

Recent developments around Foundation Large Language Models (LLMs) \cite{brown2020language,touvron2023llama,achiam2023gpt,team2023gemini,jiang2024mixtral, team2024gemma,bai2023qwen,cai2024internlm2,bi2024deepseek,kallappa2025krutrim} and \textit{Visual instruction tuning}~\cite{liu2024visual,liu2024improved} have significantly advanced Vision Language Models (VLMs)~\cite{bai2023qwenvl,chen2024far,lu2024deepseek,beyer2024paligemma,wang2023cogvlm,chen2022pali,laurenccon2024matters,tong2024cambrian,abdin2024phi,xue2024xgen,khan2025chitrarth}, enabling seamless multimodal processing of visual and linguistic data. Much of the success of these models could be attributed to the availability of the large amount of training datasets~\cite{Schuhmann2021LAION400MOD, sharma2018conceptual,kakaobrain2022coyo-700m, Schuhmann2022LAION5BAO, laurenccon2024matters, tong2024cambrian,rodriguez2024bigdocs}. However, most of the existing multimodal research is predominantly focused on single-image reasoning, while a recent line of work has begun addressing the complexities of multi-image scenarios ~\cite{Alayrac2022FlamingoAV, Zhu2023MultimodalCA, laurençon2023obelics, jiang2024mantis, McKinzie2024MM1MA, Chameleon,laurenccon2024matters}. A key factor in these advancements has been the use of interleaved text-image data which offers several compelling advantages: \textit{1.) Real-world applicability}, as it reflects the way humans typically process information, such as reading documents with both text and images \cite{alayrac2022flamingo,jiang2024mantis}; \textit{2.) Versatility across scenarios}, providing a unified approach to various tasks like single/multi-image, video, and 3D data \cite{li2024llavanextinterleave,li2024llavaonevision,damonlpsg2024videollama2,damonlpsg2025videollama3}; \textit{3.) State-of-the-art performance}, with models trained on interleaved data consistently outperforming those trained on image-text captioning datasets \cite{alayrac2022flamingo,laurençon2023obelics,futeral2024moscar}; \textit{4.) In-context learning (ICL)}, where interleaved formats improve the model's ability to follow instructions and adapt to multi-image settings \cite{laurençon2023obelics,futeral2024moscar}; and \textit{5.) Few-shot learning}, with recent studies demonstrating that interleaved data is crucial for achieving strong few-shot learning performance \cite{alayrac2022flamingo,mckinzie2024mm1,laurençon2023obelics}.

However, despite these advancements, overwhelming focus remains on English-centric and Western datasets, leaving many of the world's languages and diverse cultural contexts underrepresented \cite{yue2024pangea, khan2025chitrarth, nayak2024benchmarking}, particularly Indian languages.
While there have been recent efforts to develop inclusive multilingual multimodal models \cite{yue2024pangea, alam2024mayainstructionfinetunedmultilingual, maaz2024palo, khan2025chitrarth,khan2025chitranuvad}, most of these works leverage English dataset translations, failing to capture the cultural nuances \& linguistic diversity. 


To address this \textit{Language diversity gap in multimodal datasets}, we introduce \textbf{Chitrakshara} (``Chitra": Image and ``Akshara": Text) series\footnote{Dataset released at \url{https://huggingface.co/datasets/krutrim-ai-labs/Chitrakshara}}, consisting of \textit{1). Chitrakshara-IL}: a large-scale, interleaved pre-training dataset  comprising of approximately 193M images, 30B text tokens, and 50M multilingual documents sourced from Common Crawl spanning 11 languages. \textit{2). Chitrakshara-Cap}: 44M image-text pairs with 733M tokens. 
The primary objectives of Chitrakshara are to (i) support the development of vision-language models tailored for Indic languages, (ii) ensure linguistic and domain diversity in multimodal datasets, and (iii) improve the overall quality and representation of Indic languages in AI research. We outline a robust data collection pipeline, incorporating meticulous filtering and evaluation steps to maintain dataset quality, cultural relevance, and safety. Furthermore, we conduct an extensive quality and diversity analysis to assess the representativeness of various Indic languages and modalities within the dataset.
Our contributions could thus be summarized as follows:
\begin{itemize}[noitemsep]
    \item We introduce a large-scale, high-quality, India-focused, interleaved image-text dataset, \textbf{Chitrakshara-IL} for 
    training culturally inclusive VLMs.
    \item We also provide an image captioning dataset \textbf{Chitrakshara-Cap} based on corresponding descriptions for training a multilingual Vision encoder (ViT) \cite{alexey2020image}. 
    \item We outline a detailed methodology for creating a multimodal dataset from web data, including steps for data collection, filtering, cleaning, and deduplication, with specific adaptations for Indic languages.
    \item We conduct a comprehensive analysis of the dataset's characteristics, including language distribution, image properties, and domain representation, offering insights into its suitability for various multimodal learning tasks.
\end{itemize}

\begin{figure}
    \centering
\includegraphics[width=.9\linewidth]{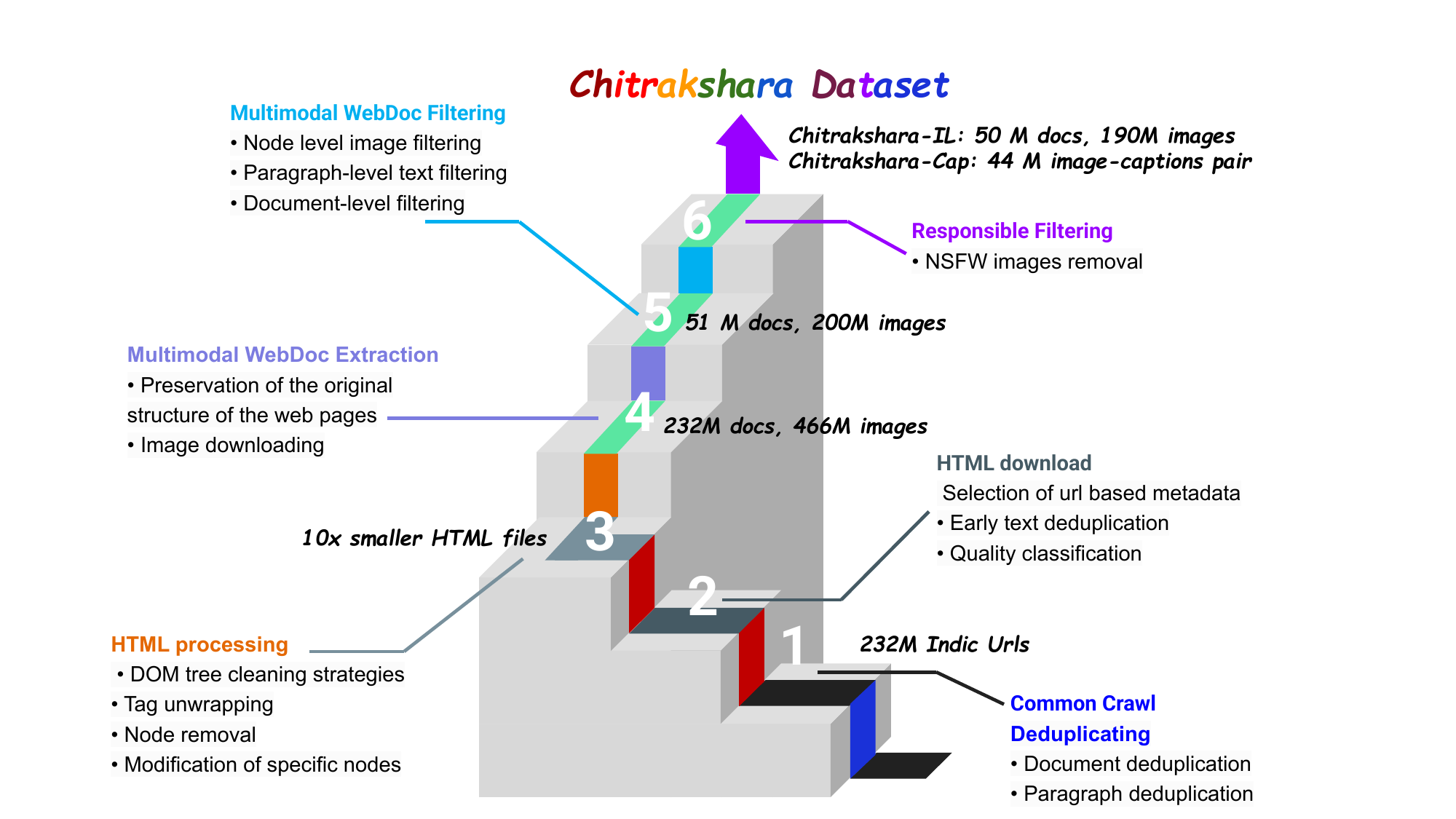} \caption{Chitrakshara dataset creation pipeline}
    \label{fig:pipeline}
\end{figure}



\begin{figure}[htbp]
    \centering
    \includegraphics[width=0.82\linewidth]{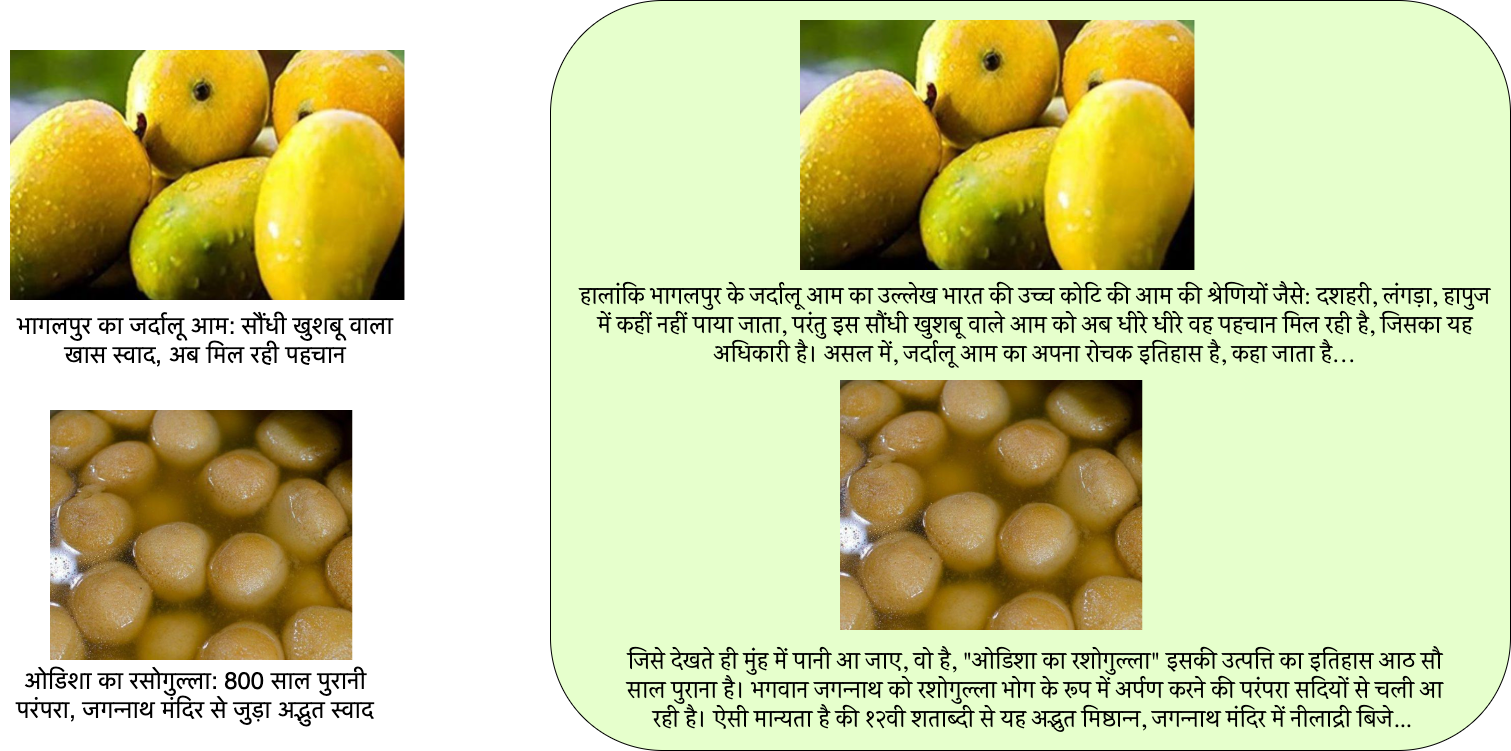}
\caption{Illustration of multimodal document extraction from the web. On the left, Chitrakshara-Cap includes image alt-text pairs, while on the right, Chitrakshara-IL retains the interleaved structure (truncated) of text \& images from the source Hindi document.}
    \label{fig:chitrakshara_teaser}
\end{figure}

\section{Related Work}
\label{sec:related-work}


\subsection{Web crawled datasets}
For text-based pretraining, large-scale datasets such as The Pile~\citep{Gao2020ThePA}, C4~\cite{raffel2020exploring}, RedPajama~\citep{together2023redpajama}, RefinedWeb~\citep{Penedo2023TheRD}, Dolma~\citep{Soldaini2024DolmaAO}, DataComp-LM~\citep{Li2024DataCompLMIS}, and FineWeb~\citep{penedo2024fineweb} have been instrumental in training LLMs. In the domain of multimodal datasets, early efforts focused on image-captioning datasets, as demonstrated by LAION-400M~\citep{Schuhmann2021LAION400MOD}, COYO-700M~\citep{kakaobrain2022coyo-700m}, ConceptualCaptions \cite{sharma2018conceptual} and LAION-5B~\citep{Schuhmann2022LAION5BAO}. 
However, most of these datasets predominantly feature English and other high-resource languages, with minimal representation of Indian languages and cultural contexts.




\subsection{Multimodal Interleaved datasets}

Recent efforts have focused on large-scale English multimodal interleaved datasets from Common Crawl to enhance reasoning abilities, including Flamingo \cite{Alayrac2022FlamingoAV}, CM3 \cite{Aghajanyan2022CM3AC}, Kosmos \cite{Huang2023LanguageIN}, and Multimodal-C4 \cite{Zhu2023MultimodalCA}, with OBELICS \cite{laurençon2023obelics} being the first large-scale open-source variant. Chameleon \cite{Chameleon} and MM1 \cite{McKinzie2024MM1MA} reported improved performance based on OBELICS type internal datasets, while MINT-1T \cite{awadalla2024mint} further expanded pretraining dataset to 1T tokens. CoMM \cite{chen2024comm} on the other hand explored other diverse data sources, and OmniCorpus \cite{li2024omnicorpus} also developed a bilingual English-Chinese dataset. Additionally, Mantis \cite{jiang2024mantis}, MIMIC-IT \cite{Li2023MIMICITMI}, and Multimodal ArXiv \cite{Li2024MultimodalAA} constructed instruction-tuning datasets using interleaved text-image data. Closely related, mOSCAR \cite{futeral2024moscar} created a multilingual interleaved dataset for 163 languages in parallel to our work, though its primary focus remains on European languages, leading to lower quality for Indic and other low-resource languages. We provide a comparative analysis and survey of these datasets in Table \ref{tab:dataset-comparison} (Appendix).

\subsection{India-centric multilingual datasets}

Relatively few efforts have been made to develop large-scale language models specifically for Indian languages. Some initiatives extended and fine-tuned English-centric models ~\citep{gala2024airavata,kohli2023building,sarvam2023openhathi,balachandran2023tamilllama,nanda2024choudhury}, while there remains a few exceptions trained from scratch~\citep{kallappa2025krutrim,bendale2024sutra,sarvam2024llm}. In parallel, a few multilingual datasets have been developed to enhance Indic language model training. IndicNLP corpora~\citep{kunchukuttan2020indicnlpcorpus} and IndicCorp~\citep{kakwani-etal-2020-indicnlpsuite} aggregated web-based content to create datasets spanning multiple Indian languages. More recently, Sangraha~\citep{khan2024indicllmsuite} introduced a large-scale corpus with 251B tokens covering 22 languages. However, these efforts predominantly focus on textual data rather than 
multimodal resources in contrast to our work. 
\begin{table*}[ht]
\centering
\resizebox{\textwidth}{!}{
\begin{tabular}{cccccccccccc}
\hline
\textbf{Hindi} & \textbf{Bengali} & \textbf{Tamil} & \textbf{Malayalam} & \textbf{Telugu} & \textbf{Marathi} & \textbf{Kannada} & \textbf{Gujarati} & \textbf{Punjabi} & \textbf{Oriya} & \textbf{Assamese} & \textbf{Total}\\
\hline
90M & 55M & 28M & 14M & 12M & 11M & 7.5M & 6.5M & 3.4M & 2.3M & 0.76M & 230M\\
\hline
\end{tabular}
}
\caption{Initial distribution of URLs from Common Crawl after deduplication.}
\label{tab:url-distribution}
\end{table*}

\section{Dataset: Chitrakshara}
\label{sec:dataset}

Our multi-lingual data creation pipeline for Chitrakshara-IL in Figure \ref{fig:pipeline} is heavily borrowed from English-only OBELICS \cite{laurençon2023obelics}, which extracts interleaved multimodal documents from CommonCrawl's (CC) \cite{commoncrawl2007}
Web ARchive Content (WARC) files. Figure \ref{fig:chitrakshara_teaser} shows an example document, more in Appendix. In addition, we extend the pipeline to also create Chitrakshara-Cap consisting of image and alt-text pairs\footnote{Alt text, or alternative text is a short description of an image on a web page, commonly used to create web-crawled captioning datasets.}, discussed in the following sections.





\subsection{HTML Pipeline}


Given that CC contains approximately 50B web pages\footnote{\url{https://registry.opendata.aws/commoncrawl/}}, with English dominating around 46\% of documents, Indian languages remain significantly underrepresented\footnote{\url{https://commoncrawl.github.io/cc-crawl-statistics/plots/languages}}, around 1\% \cite{kallappa2025krutrim}. For instance, Malayalam constitutes only 0.017\% of a specific crawl's records\footnote{\url{https://blog.qburst.com/2020/07/extracting-data-from-common-crawl-dataset/}}, while Hindi—despite being the third most spoken language globally—contributes merely 0.2\% of CC data. We thus use 95 CC dumps spanning from years 2013 to 2023 to maximize document coverage and curate a multimodal dataset over 230 million URLs, filtering Indic language web documents using FastText LID (language detector) \cite{joulin2016fasttext} and other deduplication heuristics on CC data. Table \ref{tab:url-distribution} provides the language distribution of the considered URLs in the corresponding WARC files. 



\subsection{Content Refinement Pipeline}



Next, we develop a rule-based DOM (Document Object Model) pruning framework to remove extraneous elements from HTML documents. Leveraging prior research \cite{laurençon2023obelics}, we extract text from specific HTML tags called DOM text nodes (eg. \texttt{<p>}, \texttt{<h*>}, and \texttt{<title>}, etc.) and \texttt{<img>} tags as DOM image nodes. We apply context-aware rules to eliminate unnecessary elements while preserving key structural components.
Our approach also involves converting formatting tags into standard line breaks, condensing redundant whitespace, and removing HTML comments. These refinements resulted in a tenfold reduction in HTML size while maintaining 98\% of the essential text and images. 



\subsection{Multimodal Document Assembly}

Once the HTML documents were cleaned, they were transformed into structured multimodal documents while maintaining their original layout semantics. This conversion process involved linearizing nested DOM structures into interleaved text-image sequences. We follow OBELICS in meticulously preserving the document’s original structure by retaining line breaks, paragraph boundaries, and layout separators. Image elements were extracted alongside their contextual descriptions to maintain semantic coherence. 

\begin{table*}[htbp]
    \centering
    \resizebox{0.99\linewidth}{!}{%
    \begin{tabular}{lrrrrrrrrrr}
        \hline
        \multirow{2}{*}{\textbf{Language}} & \multicolumn{2}{c}{\textbf{Documents}} & \multicolumn{2}{c}{\textbf{Tokens}} & \multicolumn{2}{c}{\textbf{Avg Tokens/Doc}} & \multicolumn{2}{c}{\textbf{Images}} & \multicolumn{2}{c}{\textbf{Avg Images/Doc}} \\
        \cline{2-3} \cline{4-5} \cline{6-7} \cline{8-9} \cline{10-11}
        & \textbf{mOSCAR} & \textbf{Chitrakshar} & \textbf{mOSCAR} & \textbf{Chitrakshar} & \textbf{mOSCAR} & \textbf{Chitrakshar} & \textbf{mOSCAR} & \textbf{Chitrakshar} & \textbf{mOSCAR} & \textbf{Chitrakshar} \\
        \hline
        Assamese  & 3.9K   & 172.9K  & 640K    & 92.9M   & 162.2  & 537.7  & 9.2K   & 559.4K  & 2.33   & 3.24 \\
        Punjabi   & 11.5K  & 481K    & 1.89M   & 284.1M  & 164.2  & 591.3  & 46.2K  & 1.91M   & 4.02   & 3.97 \\
        Odia      & 4.3K   & 601K    & 379K    & 331.6M  & 87.7   & 551.9  & 15.6K  & 2.87M   & 3.61   & 4.78 \\
        Gujarati  & 23.1K  & 1.12M   & 3.32M   & 662.4M  & 144.2  & 590.7  & 91.3K  & 3.62M   & 3.96   & 3.23 \\
        Kannada   & 13.0K  & 1.50M   & 1.44M   & 864.6M  & 111.2  & 575.3  & 42.6K  & 4.95M   & 3.28   & 3.30 \\
        Telugu    & 23.0K  & 1.98M   & 2.27M   & 1.16B   & 99.0   & 586.1  & 81.0K  & 6.27M   & 3.53   & 3.17 \\
        Marathi   & 50.4K  & 3.14M   & 6.69M   & 1.82B   & 132.7  & 579.0  & 164K   & 10.96M  & 3.25   & 3.49 \\
        Malayalam & 14.1K  & 3.33M   & 1.69M   & 1.97B   & 119.4  & 589.7  & 52.7K  & 12.05M  & 3.73   & 3.62 \\
        Bengali   & 270.4K & 6.06M   & 35.9M   & 2.93B   & 132.6  & 484.3  & 947K   & 27.6M   & 3.50   & 4.55 \\
        Tamil     & 36.2K  & 6.69M   & 4.83M   & 4.13B   & 133.6  & 617.5  & 168K   & 23.39M  & 4.64   & 3.49 \\
        Hindi     & 579.4K & 25.4M   & 122.6M  & 14.9B   & 211.5  & 586.9  & 1.83M  & 99.3M   & 3.16   & 3.91 \\
        \hline
    \end{tabular}%
    }
    \caption{Comparison of Chitrakshara-IL and mOSCAR, the only other interleaved dataset supporting Indian languages.}
    \label{tab:moscar-comparison}
\end{table*}

\begin{table}[h]
    \centering
    \resizebox{0.85\linewidth}{!}{

    \begin{tabular}{lrrr}
        \hline
        \textbf{Language}  & \textbf{\# Pairs} & \textbf{\# Tokens} &  \textbf{\# Avg. tokens} \\
        \hline
        Punjabi        & 127.8K  & 2.49M  & 19.46  \\
        Assamese       & 132.8K  & 2.57M  & 19.34 \\
        Kannada        & 475.3K  & 9.05M  & 19.05 \\
        Gujarati       & 525.1K  & 11.62M & 22.12 \\
        Telugu         & 866.7K  & 17.96M & 20.53 \\
        Malayalam      & 1.16M   & 23.54M & 20.38  \\
        Odia           & 628.3K  & 8.61M  & 13.71  \\
        Marathi        & 1.87M   & 28.73M & 15.33  \\
        Tamil          & 2.49M   & 45.68M & 18.33 \\
        Bengali        & 3.42M   & 56.18M & 16.43 \\
        English        & 11.29M  & 148.35M & 13.13  \\
        Hindi          & 21.29M  & 379.23M & 17.81 \\
        \hline
    \end{tabular}
    }
    \caption{Language distribution for Chitrakshar-Cap dataset.}
    \label{tab:cap_stats}
\end{table}

\subsection{Hierarchical Content Filtering}
To ensure dataset quality, we further implemented a multi-stage filtering framework.

\textbf{Image Filtering:}
 At the node level, images were discarded if they did not meet predefined criteria, such as format (restricted to JPEG, PNG, and WEBP), dimensions (at least 150 pixels on either side), or aspect ratio constraints (between 1:5 and 5:1). 
 
 
 \textbf{Paragraph level Filtering:}
 Similarly, textual content was filtered using linguistic heuristics adapted from existing research \cite{kallappa2025krutrim}. Specifically, paragraphs with fewer than 8 words were discarded. 
 
\textbf{Document-Level Validation:}
We also conducted holistic evaluations at the document level to determine the retention or exclusion of an entire webpage. Particularly, we enforced multimodal balance by rejecting documents that contain no images or more than 30 images. We also apply coherence checks to remove pages with repetitive patterns.

\subsection{Additional Heuristics}

In addition to the filtering techniques above, we develop rule-based heuristics to eliminate ``Continue Reading" links, publication dates, social media sharing prompts, ``About Us" sections, and other metadata including navigation-related text such as scroll and pause instructions, notifications, subscription prompts, and alerts. To avoid irrelevant images, we remove images with filename containing substrings like ``default" or ``placeholder" or alt text containing block words. 
To identify inappropriate or NSFW images we check if either the filename or alt text contains NSFW words
as a substring. If so, we then remove the entire document containing that image.

We thus generate Chitrakshara-IL with 193M images (53M docs) using a unified pipeline. Applying additional filtering, we pair images with metadata alt-text (distinct from document content), to form Chitrakshara-Cap. Notably, alt-text may remain in English even when the document is in another language. We provide more specific details for each component of our pipeline as well as the implementation and infrastructure in Appendix (Section \ref{appendix:pipeline}).






\section{Analysis}
\label{sec:analysis}

Tables \ref{tab:moscar-comparison} and \ref{tab:cap_stats} show the language-wise distribution of Chitrakshara-IL and Chitrakshara-Cap datasets respectively. 
Additionally, we compare key statistics with mOSCAR, the only other multilingual interleaved dataset that covers 163 languages, including the Indian languages examined in our study.
Our findings indicate that our dataset contains significantly more documents, tokens, and images while also features a higher average of tokens and images per document across most Indian languages. For example, Hindi has 25M documents versus mOSCAR’s 579K.
We attribute this difference to our data collection strategy, which incorporates 95 CC dumps spanning a decade, unlike mOSCAR’s three dumps from 2023, offering a broader and more temporally diverse corpus. Additionally, mOSCAR’s filtering methods, optimized for English and European languages, may not effectively capture Indian languages. Notably, perplexity filtering (based on models trained on English corpora), as employed in OBELICS, disproportionately removes Indo-Aryan and Dravidian language content. Our approach with a focus on India languages  employs tailored filtering, ensuring better content representation. 

\begin{figure}[htbp]
    \centering    \includegraphics[width=0.8\linewidth]{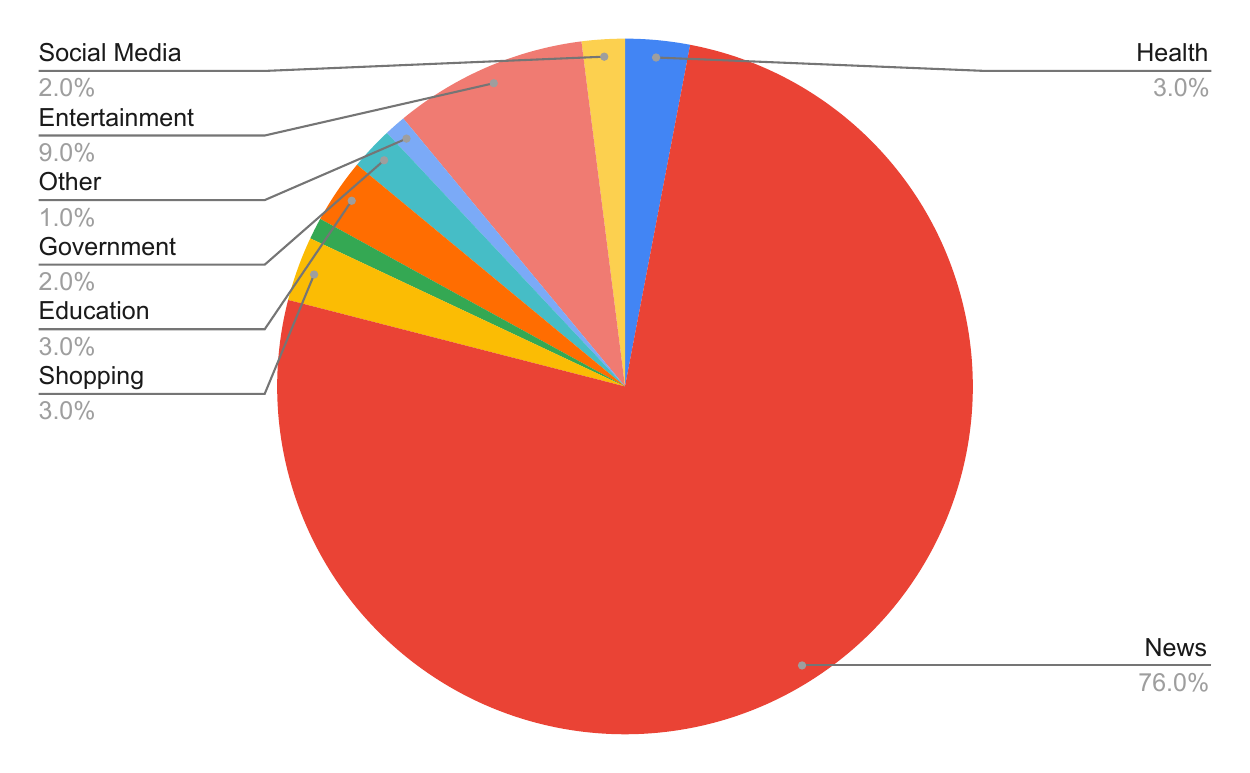}
    \vspace{-3mm}
    \caption{Domain distribution of the Chitrakshara-IL data.}
    \label{fig:domain}
\end{figure}
\begin{figure}[htbp]
    \centering
    \includegraphics[width=0.78\linewidth]{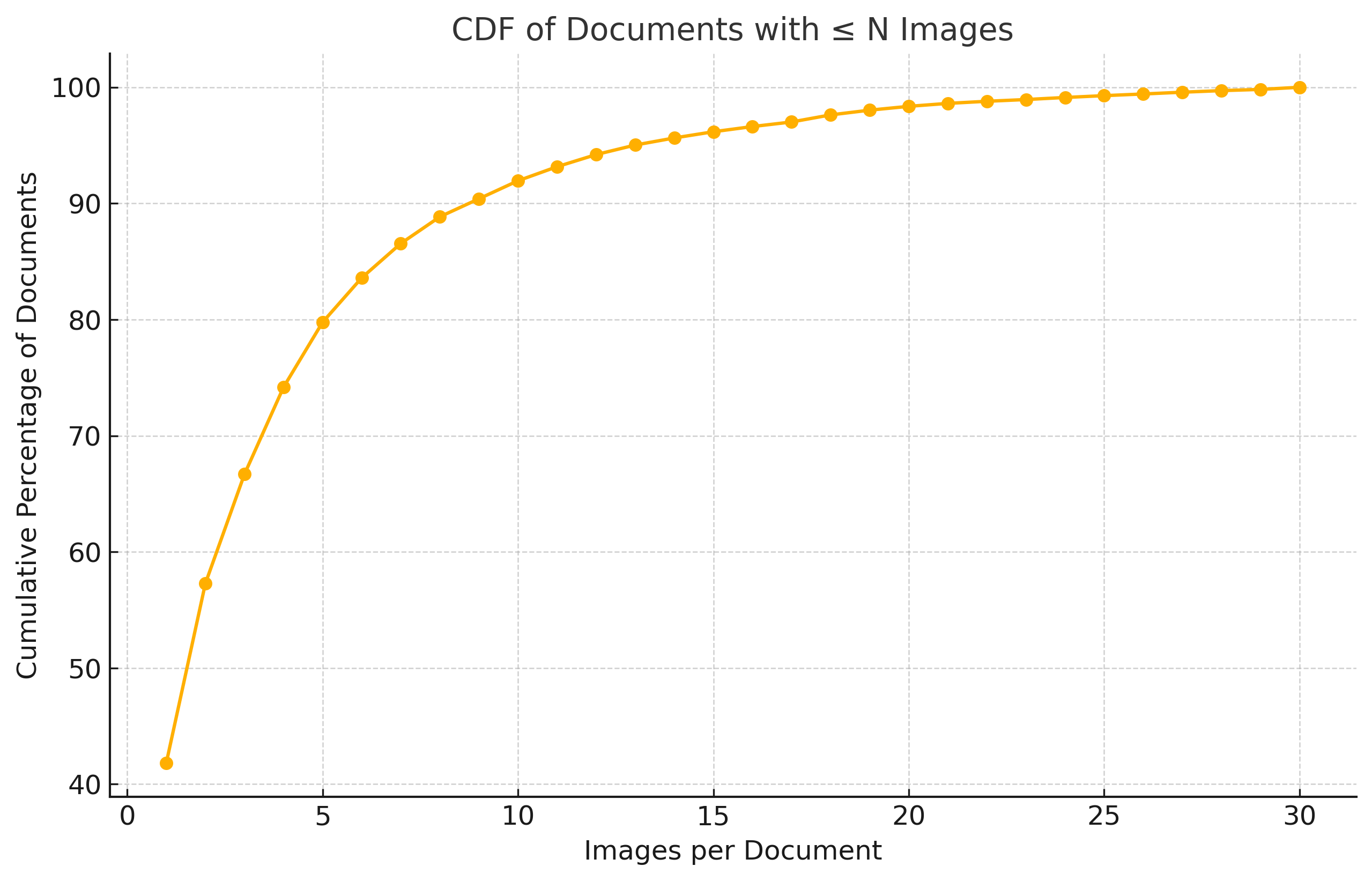}
    \vspace{-2mm}
    \caption{Cumulative image count distribution per document.}
    \label{fig:image_count}
\end{figure}

Furthermore, domain analysis (Figure \ref{fig:domain}) reveals news websites dominate (76\%) followed by entertainment (9\%), health (3\%), education (3\%), etc. mirroring Indic Sangraha \cite{khan2024indicllmsuite} and English-only interleaved OBELICS. 
We also list the top 20 domains per theme and the top 100 by document count in Figure \ref{fig:top_20_domain} and Table \ref{tab:top_domains} respectively (in Appendix). 
80\% of the interleaved documents have fewer than five images (c.f. Figure \ref{fig:image_count}). 
 Temporal distribution analysis shows most data originates from the past seven years while image size distribution indicates most images are $\sim$256 pixels per side (see Appendix Figure \ref{fig:documents_over_years} and \ref{fig:image-size}).  Lastly, we discuss the top topics across different languages (Section \ref{sec:topic_modelling} in Appendix), underscoring the diverse range of captured content.

\section{Conclusion}
\label{sec:conclusion}

We introduce the Chitrakshara dataset series, a large-scale, multilingual, and multimodal resource covering 11 Indian languages. It includes Chitrakshara-IL, an interleaved dataset with 193M images, 30B tokens, and 50M documents, and Chitrakshara-Cap, with 44M image-text pairs \& 733M tokens. Our work details the data collection, filtering, and processing pipeline, ensuring quality and diversity. By filling gaps in existing multilingual datasets, Chitrakshara facilitates the development of more culturally inclusive VLMs. Future work involves training a multilingual ViT and an interleaved VLM to evaluate its effectiveness.

{
    \small
    \bibliographystyle{ieeenat_fullname}
    \bibliography{main,mint_references}
}


\newpage


\clearpage
\setcounter{page}{1}
\maketitlesupplementary



\section*{Acknowledgements}
We thank Bhavish Aggarwal and the rest of the Krutrim team which helped with data collection at various stages. Our efforts received generous support from Krutrim cloud using Krutrim credits. We also thank the reviewers for their valuable feedback and suggestions. 

\section*{Author Statement}
\label{sec:license}

We acknowledge that Chitrakshara may reflect inherent biases present in online content, as the dataset is sourced from the internet. Nonetheless, its multilingual and inclusive composition marks a significant step toward enhancing the accessibility of diverse languages, cultures, and communities for training India-centric Vision-Language Models (VLMs). While we have taken rigorous measures to ensure the accuracy and legality of the dataset, we cannot guarantee its absolute completeness or correctness. Consequently, the authors assume no liability for any potential legal or ethical concerns, including but not limited to copyright infringement, privacy violations, or the misuse of sensitive information.

\begin{table}[htbp]
\centering
\resizebox{0.99\linewidth}{!}{

\begin{tabular}{lccccc}
\toprule
Dataset & \# Tokens & \# Images & \# Docs & Multilingual & Data Sources \\
\midrule
\textit{Image-text Paired Datasets} \\
COYO-700M ~\citep{kakaobrain2022coyo-700m} & 12.9B & 747M & -  & \textcolor{red}{\xmark} & CC \\
LAION-5B ~\citep{Schuhmann2022LAION5BAO} & 135B & 5B & - & \textcolor{green}{\cmark} & CC \\
\textbf{Chitrakshara-Cap} & 733M & 44M & - & \textcolor{green}{\cmark} & CC \\
\midrule
\textit{Image-text Interleaved Datasets} \\
CM3~\cite{Aghajanyan2022CM3AC} & 223B & 373M & 10.7M & \textcolor{red}{\xmark} & CC \\
Multimodal-C4~\cite{Zhu2023MultimodalCA} & 43B & 571M & 101M & \textcolor{red}{\xmark} & CC \\
OBELICS \cite{laurençon2023obelics} & 115B & 353M & 141M & \textcolor{red}{\xmark} & CC \\
CoMM \cite{chen2024comm} & 139M & 2.28M & 227K & \textcolor{red}{\xmark} & Curated \\

MINT-1T \cite{awadalla2024mint} & 1.02T & 3.42B & 1.05B & \textcolor{red}{\xmark} & CC, PDFs, ArXiv \\
OmniCorpus \cite{li2024omnicorpus} & 1.7T & 8.6B & 2.2B & \textcolor{red}{\cmark} & CC, CW, YT \\

mOSCAR \cite{futeral2024moscar} & 214B & 1.2B & 315M & \textcolor{green}{\cmark} & CC \\
\textbf{Chitrakshara-IL} & 30B & 193M & 50M & \textcolor{green}{\cmark} & CC \\
\bottomrule
\end{tabular}
}
\caption{\textbf{Survey of multimodal datasets:} Chitrakshara-IL represents interleaved dataset while Chitrakshara-Cap represents alt-text image pairs. CC represents data is sourced from Common Crawl.  CoMM followed a different recipe of using curated sources consisting of WikiHow, eHow, Story bird, StoryGen, Instructables against using CC dumps. Omnicorpus is bilingual supporting Chinese and English only  sourced also from YouTube (YT) and other Chinese websites (CW) apart from CommonCrawl. mOSCAR is the only multi-lingual multimodal interleaved dataset that also supports Indian languages but it's focus remain primarily on Western languages.}
\label{tab:dataset-comparison}
\end{table}


\section{Data pipeline technical insights}
\label{appendix:pipeline}

We implemented our code in python building upon the OBELICS\footnote{\url{https://github.com/huggingface/OBELICS}} framework, adapting it for websites with rich content from Indian languages. 

\subsection{HTML Pipeline}

We begin by gathering 95 Common Crawl dumps spanning the years 2013 to 2023. Unlike projects such as RedPajama \cite{together2023redpajama}, which construct large-scale datasets using only five minimally overlapping dumps\footnote{\url{https://commoncrawl.github.io/cc-crawl-statistics/plots/crawloverlap}}, our approach involves a more extensive collection. This allows us to include a broader range of Indian documents, which otherwise account for just 1\% of the data.

One of the primary challenges we faced in this step was overcoming HTTP rate limits, which frequently led to throttling during direct access to Common Crawl’s Meta or WARC files via HTTP. To mitigate this, we optimized data retrieval by implementing a distributed query system. 
To optimize data transfer, we developed an S3-to-S3 pipeline, which allowed us to migrate 24 terabytes of filtered metadata directly between Amazon S3 buckets, eliminating HTTP bottlenecks and achieving sustained transfer rates of 10 Gbps. As a result, we successfully processed the metadata for 230 million URLs in under 24 hours using AWS infrastructure, reducing computational costs by 60\% compared to traditional single-node scraping approaches.

\subsubsection{WARC Retrieval and Distributed Processing}

One major limitation was parallelization. Performance degradation occurred due to resource contention when exceeding a threshold of concurrent processes. Additionally, unpredictable network latencies led to idle compute resources, further slowing the process. To overcome this, we implemented an adaptive parallelization strategy where network-bound tasks, such as WARC downloads, were structured to overlap with computation. By directly transferring data to AWS S3 instead of writing it to disk, we significantly reduced I/O overhead.

Automation and orchestration played a key role in optimizing workflow. Using a bash-based job management system, we automated process distribution, implemented retries for failed downloads, and consolidated output using dynamic job queues. Ansible playbooks were used to synchronize configurations across all 25 nodes, ensuring a consistent environment. These measures resulted in a 30\% speedup in processing time while reducing idle node time by 10\%, allowing us to process the dataset in under two days. We also use a modified version of readability-lxml\footnote{\url{https://github.com/buriy/python-readability}} library to extract the primary text from web pages.


\subsection{Content Refinement Pipeline}


To refine raw HTML documents and remove irrelevant elements such as advertisements and template-based components, we develop a rule-based DOM pruning framework. 
We extract text from specific HTML tags that typically contain the primary content of web pages, referred to as DOM text nodes (\texttt{<p>, <h*>, <title>}, etc.) and all \texttt{<img>} tags, as DOM image nodes.
We implement context-aware rules by defining cascading filters to remove nodes matching spam indicators (e.g., class="advert", excessive \texttt{<script>} density) while preserving semantic containers. Using the selectolax\footnote{\url{https://github.com/rushter/selectolax}} library for efficient HTML parsing, we applied these rules to eliminate unnecessary elements while preserving key structural components. Our approach involves converting formatting tags (e.g., \texttt{<br>}) into standard line breaks, condensing redundant whitespace, and removing HTML comments. Additionally, recursive cleaning operations unwrapped unnecessary styling elements (e.g., \texttt{<i>}, \texttt{<span>}) and streamlined the DOM hierarchy by collapsing redundant nodes. These refinements resulted in a tenfold reduction in HTML size while maintaining 98\% of the essential text and images. We follow similar strategy as OBELICS in unwrapping the style element tags.

To further enhance document quality, we implement a systematic filtering strategy to retain only structurally and semantically relevant tags. Tags critical for document structure (e.g., \texttt{<p>}, \texttt{<h1>}–\texttt{<h6>}, \texttt{<section>}) and media representation (e.g., \texttt{<img>}, \texttt{<video>}, \texttt{<figure>}) were preserved, while those associated with navigation menus, headers, and footers were removed. Specific \texttt{<div>} elements containing identifiers such as \texttt{footer}, \texttt{navbar}, or \texttt{menu} were also discarded to eliminate noisy content. Additionally, nodes with the class \texttt{more-link}, which often signaled content transitions, were replaced with a placeholder token (\texttt{END\_OF\_DOCUMENT\_TOKEN\_TO\_BE\_REPLACED}) similar to OBELICS pipeline. These preprocessing techniques ensured a cleaner and more structured dataset, significantly optimizing the extraction of textual and visual elements for downstream applications.

\subsection{Multimodal Document Assembly}

HTML documents that were cleaned in the previous step were transformed into structured multimodal documents while maintaining their original layout semantics. This conversion process involved linearizing nested DOM structures into interleaved text-image sequences, embedding structural markers such as \texttt{<SECTION>} and \texttt{<FIGURE>} to ensure proper content delineation. We meticulously preserve the document’s original structure by retaining line breaks, paragraph boundaries, and layout separators. Image elements were extracted alongside their contextual descriptions, such as \texttt{<figcaption>} tags, to maintain semantic coherence. To facilitate large-scale image retrieval, we employed the \texttt{img2dataset} \cite{beaumont2021img2dataset} library and distributed the downloading process across 40 virtual machines. 
With a parallelized download of 3.6B image links, we achieved 55\% retrieval success, i.e.\ around 2B images. 

\subsection{Hierarchical Content Filtering}
Here we implemented multiple filtering techniques:

\textbf{Image Filtering:}
 Images at the node level were discarded if they did not meet predefined criteria, such as format (restricted to JPG, JPEG, PNG, and WEBP), dimensions (between 150 pixels on either side), or aspect ratio constraints (between 1:5 and 5:1). Additional heuristics were applied to remove generic and low-value images by detecting substrings such as \texttt{logo}, \texttt{icon}, \texttt{banner}, \texttt{social}, and \texttt{widget} in URLs. 
 
 \textbf{Paragraph Filtering:}
 Similarly, textual content was filtered using linguistic heuristics adapted from existing research \cite{kallappa2025krutrim}. Paragraphs with fewer than eight words were discarded. We also ensure stopword density remained above 5\% to filter out machine-generated lists and incoherent content. Table \ref{tab:cutoffs_text_filters} presents the filters that were used. 
 
\textbf{Document-Level Validation:}
At the document level, we enforced multimodal balance by rejecting documents containing no or more than 30 images. Additionally, coherence checks were applied to remove pages with repetitive patterns indicative of machine-generated text. Beyond node-level filtering, we conducted holistic evaluations at the document level to determine the retention or exclusion of an entire webpage. Tags associated with website navigation (\texttt{header}, \texttt{menu}, \texttt{navbar}) and footer sections were removed, and transitional elements (\texttt{more-link}) were replaced with an end-of-document token (\texttt{END\_OF\_DOCUMENT\_TOKEN\_TO\_BE\_REPLACED}). By systematically applying these refinements, we ensured that our dataset remained both high-quality and representative of real-world multimodal web documents.




\subsection{Additional processing}
\label{sec:post-processing}
\subsubsection{Text-based filtering}
To ensure Chitrakshara is suitable for training vision-language models on interleaved image-text conversations, extensive text filtering is applied. Irrelevant elements such as “Continue Reading” links, publication dates, social media prompts, ``About Us" sections, and other metadata are removed. Additionally, navigation-related text, alerts, and subscription prompts are filtered out, keeping the dataset focused on meaningful dialogue. We explore two filtering strategies for Indian languages. The first is heuristic-based, where paragraphs are split into lines, English text is detected and removed, predefined unwanted phrases are filtered out, and short paragraphs below a word threshold are discarded. The second strategy leverages large language models (LLMs) for content filtering, but due to computational costs, we adopt the first approach. In this process, paragraphs are split at newline characters, and lines containing only English text, numbers, or symbols are removed. The filtered lines are then recombined, preserving meaningful multilingual content while eliminating noise.

\subsubsection{Image-based filtering}

We also applied filtering techniques to remove irrelevant or inappropriate images. Entries with filename containing substrings like ``default" or ``placeholder" were removed, as these typically represent empty or placeholder images that do not contribute meaningful visual content. Similarly, images containing any word among ``download", ``pdf", ``mp4", ``mp3", ``chapter", ``video", ``audio" were removed because these images did not contribute to good quality interleaved content. Regarding inappropriate or NSFW images, any image with either the filename or alt text containing NSFW words like ``s**", ``p***", ``f***" or similar words in Indian languages as a substring was identified as an NSFW image and the document containing that image was removed. This step helps eliminate non-informative images, explicit content, and media-related placeholders, ensuring that only relevant images are retained for training. 

\subsubsection{Chitrakshara-Cap filtering}
For generating Chitrakshara-Cap, we apply additional filtering of minimum 5 words in the corresponding alt-text. This was done to ensure that we get only images with corresponding relevant descriptions. We also assess the image quality by classifying images based on predefined resolution criteria: Low Resolution images have either a width or height of less than 200 pixels, High Resolution images have both width and height greater than 600 pixels, and all others fall under Mid Resolution. Analysis of the dataset revealed that 22.5\% of the images are Low Resolution, 64.1\% are Mid Resolution, and 13.4\% are High Resolution. This distribution indicates that most images in the dataset are of usable quality, with a significant proportion meeting medium and high-resolution standards.

\subsection{Infrastructure}\label{Infrastructure}

For data processing, we utilize a cluster of 25 machines with a total of 5,120 CPU cores, consisting of 15 high-performance nodes (256 cores, 512 GB RAM) and 10 mid-range nodes (128 cores, 256 GB RAM). The dataset was processed in 900 batches. We empirically download 40 images using multi-processing, which provided us the relevant speedup as well as the best download success rate. 


\begin{algorithm}
    \caption{Multimodal Dataset Creation Pipeline}
    \label{alg:dataset_pipeline}
    \begin{algorithmic}[1]
        \State \textbf{Input:} Common Crawl WARC files
        \State \textbf{Output:} Curated Multimodal Dataset
        
        \Procedure{DatasetCreation}{WARC\_Files}
            \State \textbf{Step 1: Identify Indic Language Web Content}
            \State Collect 95 Common crawl dumps. 
            \State Identify 230M URLs related to Indian language web content.

            
            \State \textbf{Step 2: Distributed WARC Retrieval}
            \State Initialize 25-node cluster
            \State Parallelize downloads, mitigating rate limits
            \State Store extracted documents directly in AWS S3
            
            \State \textbf{Step 3: Content Refinement}
            \State Parse HTML DOM to extract meaningful content
            \State Prune unwanted elements (ads, sidebars, pop-ups)
            \State Apply rule-based filtering for noisy content
            
            \State \textbf{Step 4: Multimodal Document Assembly}
            \State Convert DOM structure to linearized text-image format
            \State Extract image URLs with corresponding captions
            \State Download images using geographically distributed proxies
            
            \State \textbf{Step 5: Hierarchical Content Filtering}
            \State \textbf{Granular Filtering:} Remove small and distorted images
            \State \textbf{Text Filtering:} Discard short, incoherent, or redundant text. Use LID to filter out paragraphs
            \State \textbf{Multimodal Validation:} Enforce image-to-text ratio constraints
            
            \State \textbf{Step 6: Infrastructure Utilization}
            \State Deploy cluster with 5,120 CPU cores
            \State Process dataset in 900 batches to optimize throughput
            
            \State \textbf{Step 7: Post-processing}
            \State Remove metadata (``Continue Reading'', dates, etc.)
            \State Detect and discard non-content elements using heuristics
            \State Perform NSFW filtering
        \EndProcedure
    \end{algorithmic}
\end{algorithm}


\begin{algorithm}
    \caption{Text-Based Filtering}
    \label{alg:text_filtering}
    \begin{algorithmic}[1]
        \Procedure{TextFiltering}{Paragraphs}
            \For{each paragraph in Paragraphs}
                \State Split paragraph into lines
                \For{each line in paragraph}
                    \If{line contains only English characters, special symbols, emojis etc. or line contains less than 4 words}
                        \State Remove line
                    \EndIf
                \EndFor
                \State Reassemble paragraph with filtered lines
            \EndFor
            \State \Return Cleaned Text
        \EndProcedure
    \end{algorithmic}
\end{algorithm}

\begin{algorithm}
    \caption{Image-Based Filtering}
    \label{alg:image_filtering}
    \begin{algorithmic}[1]
        \Procedure{ImageFiltering}{ImageEntries}
            \For{each image entry in ImageEntries}
                \If{Filename contains ``default'' or ``placeholder'' or Alt-text contains "download", "pdf", "mp4" etc.}
                    \State Remove image entry
                \EndIf
                \If{Alt-text or Filename contains NSFW substrings ( ``s**', ``p***'', etc.)}
                    \State Remove image entry (along with the doc containing it)
                \EndIf
            \EndFor
            \State \Return Filtered Image Set
        \EndProcedure
    \end{algorithmic}
\end{algorithm}

\begin{table*}[htbp]
\centering
\begin{tabular}{ lccc}
 \hline
 Metric & Cutoff type & Value (para-level)& Value (doc-level)

 \\
 \hline
 \rule{0pt}{12pt}Number of words &min &4 &10\\
 Number of words &max &1,000 &2,000\\
 Character repetition ratio &max &0.1 &0.1\\
 Word repetition ratio &max &0.1 &0.2\\
 Common word ratio &min &0.1 &0.1\\
 \hline
\end{tabular}
\vspace{0.5em}
%
\caption{Cutoff thresholds for text filters at paragraph and document levels. Cutoff values ("min" or "max") removes paragraphs or documents strictly below or above the threshold respectively.}
\label{tab:cutoffs_text_filters}
\end{table*}




\begin{figure}[h]
    \centering
    \includegraphics[width=\linewidth]{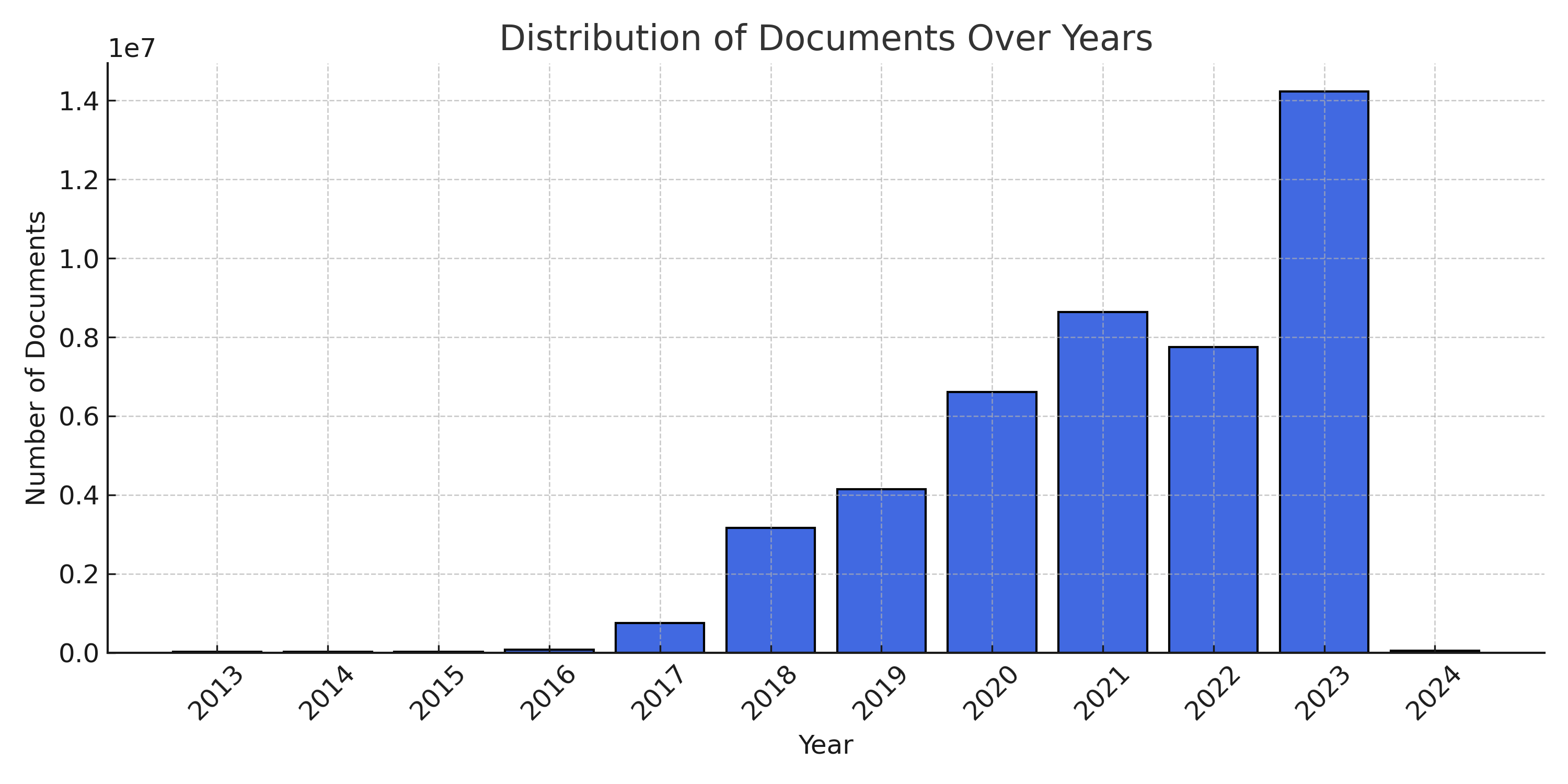}
    \caption{Distribution of Documents Over Years.}
    \label{fig:documents_over_years}
\end{figure}

\begin{figure}[htbp]
    \centering
    \includegraphics[width=\linewidth]{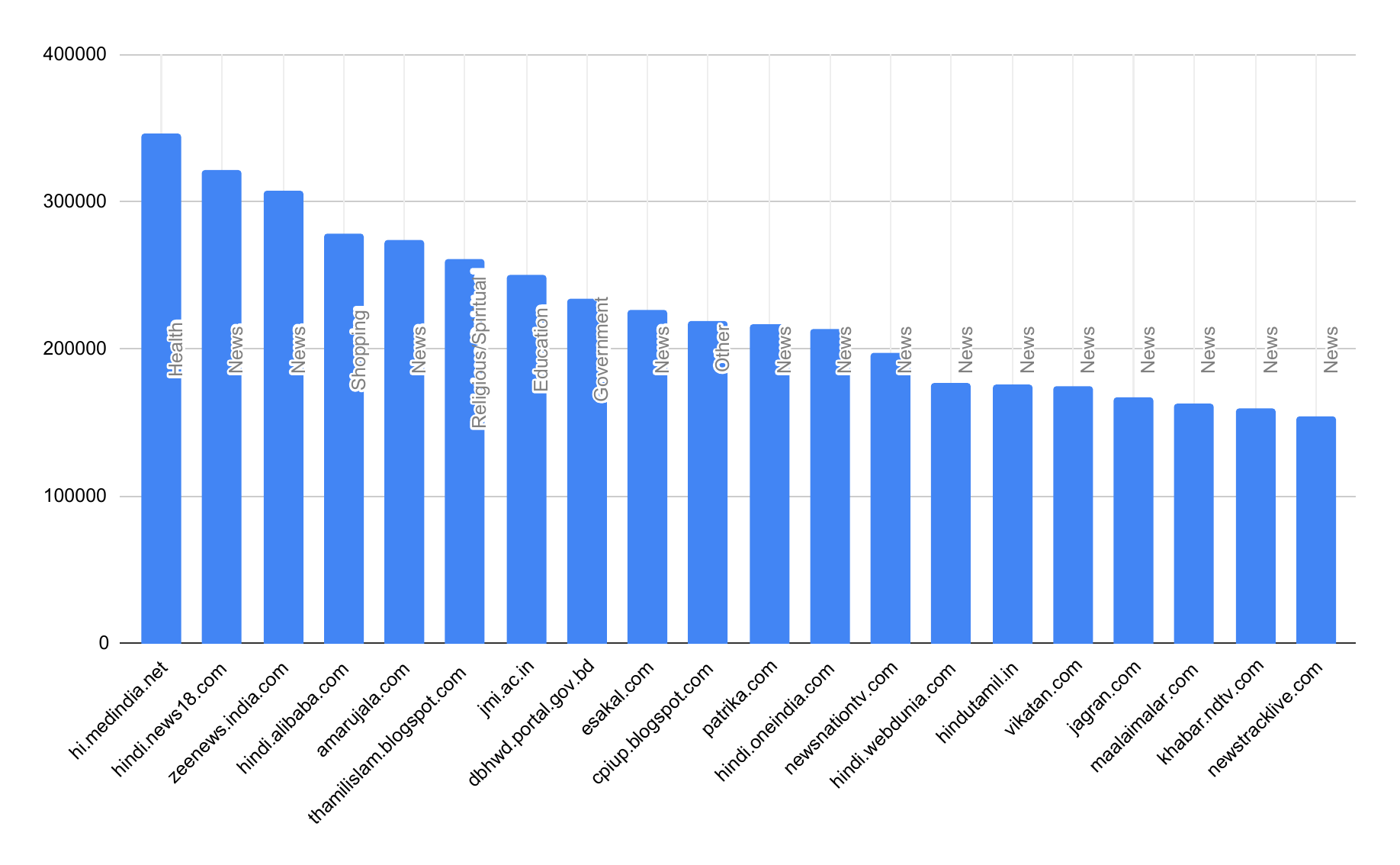}

    \caption{Top 20 Domains in Chitrakshara-IL.}
    \label{fig:top_20_domain}
\end{figure}

\begin{figure}
    \centering
\includegraphics[width=\linewidth]{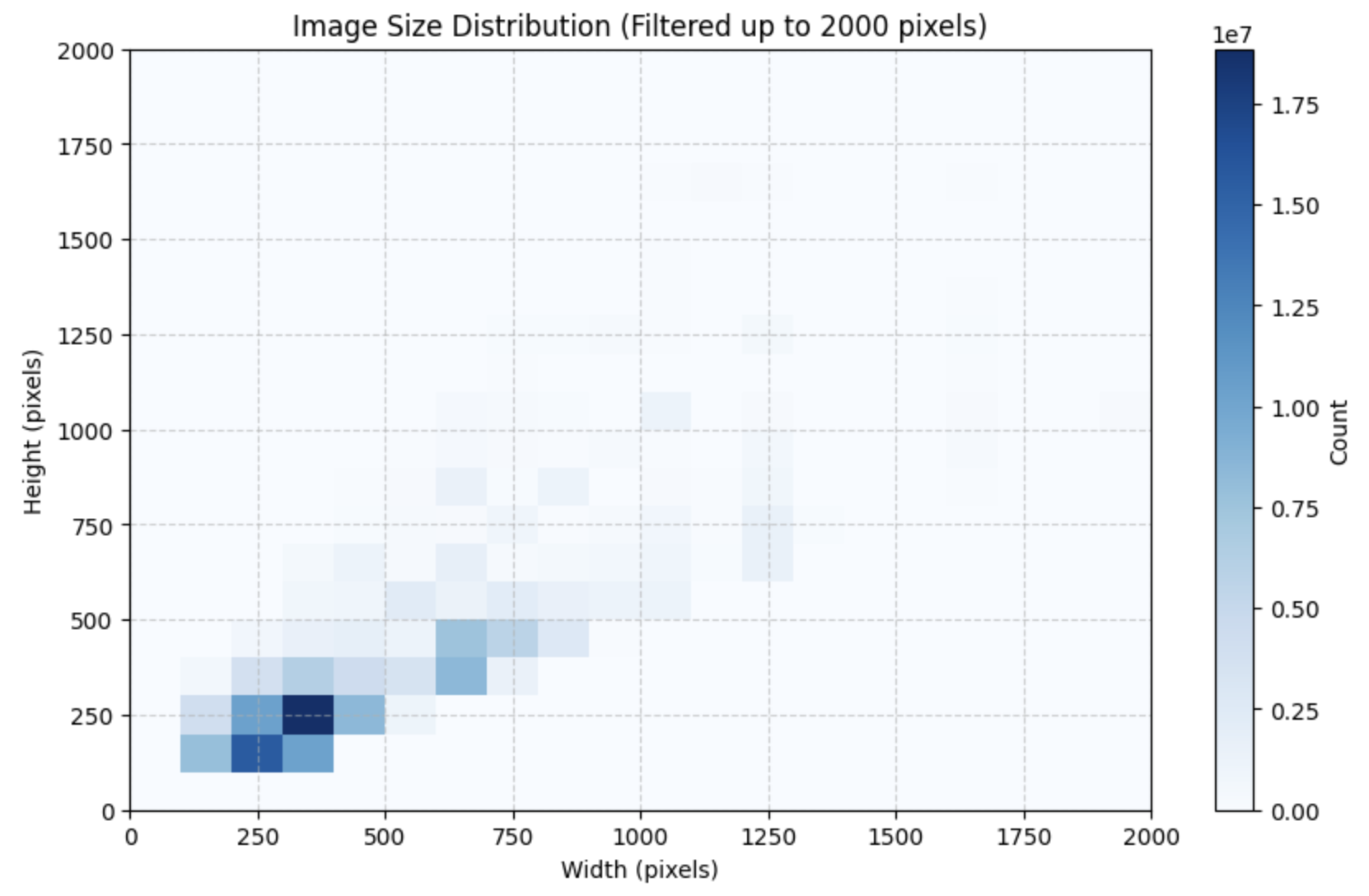} \caption{Image size distribution.}
    \label{fig:image-size}
\end{figure}

\section{Top topics across languages}
\label{sec:topic_modelling}

\begin{figure*}[htbp]  
    \centering
    \includegraphics[page=1, width=0.8\linewidth]{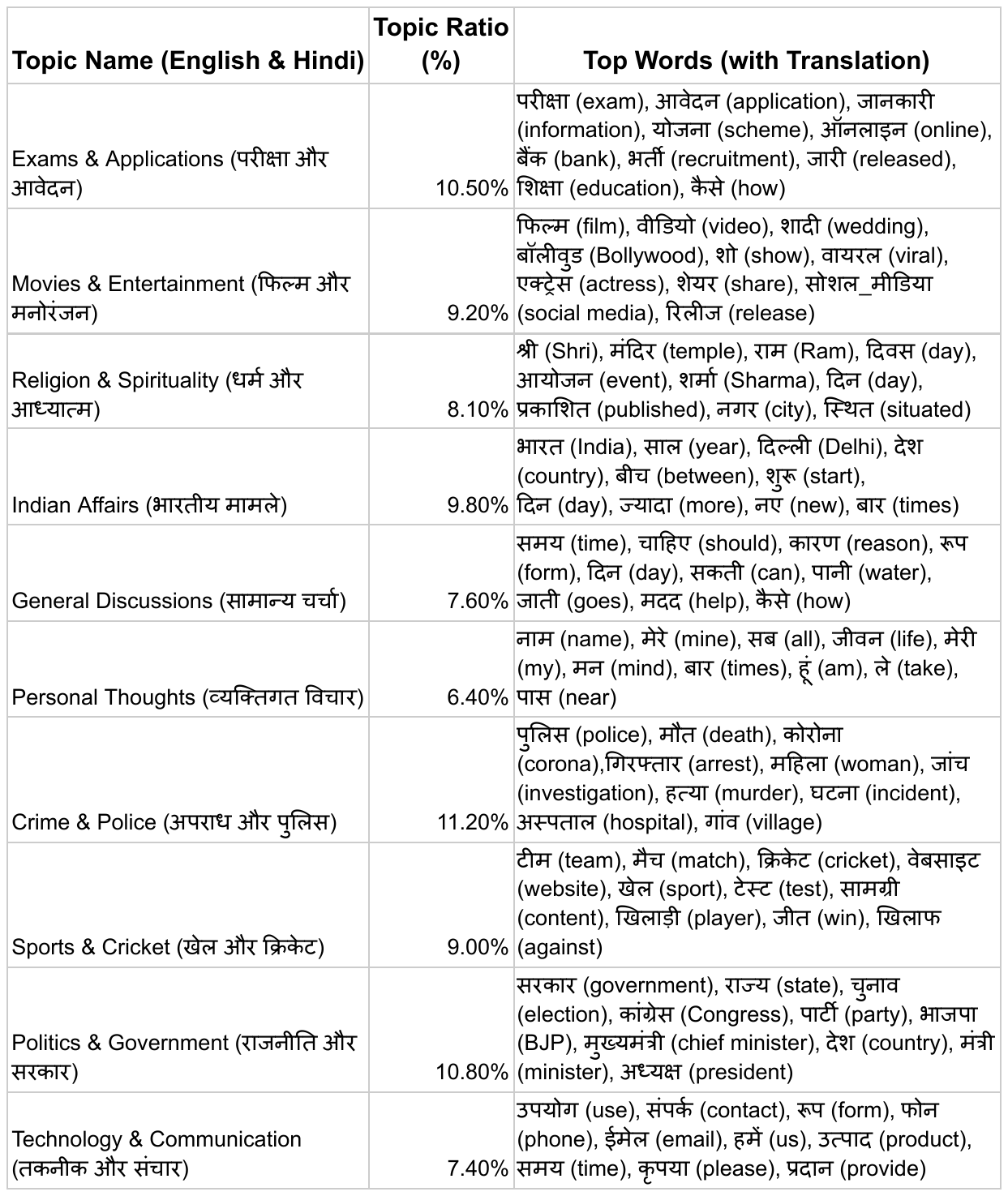}
    \vspace{-2mm}
    \caption{Topic modelling results for Hindi language in Chitrakshara-IL dataset.}
    \label{fig:topics-hindi}
\end{figure*}

\begin{figure*}[htbp]  
    \centering
    \includegraphics[page=2, width=0.8\linewidth]{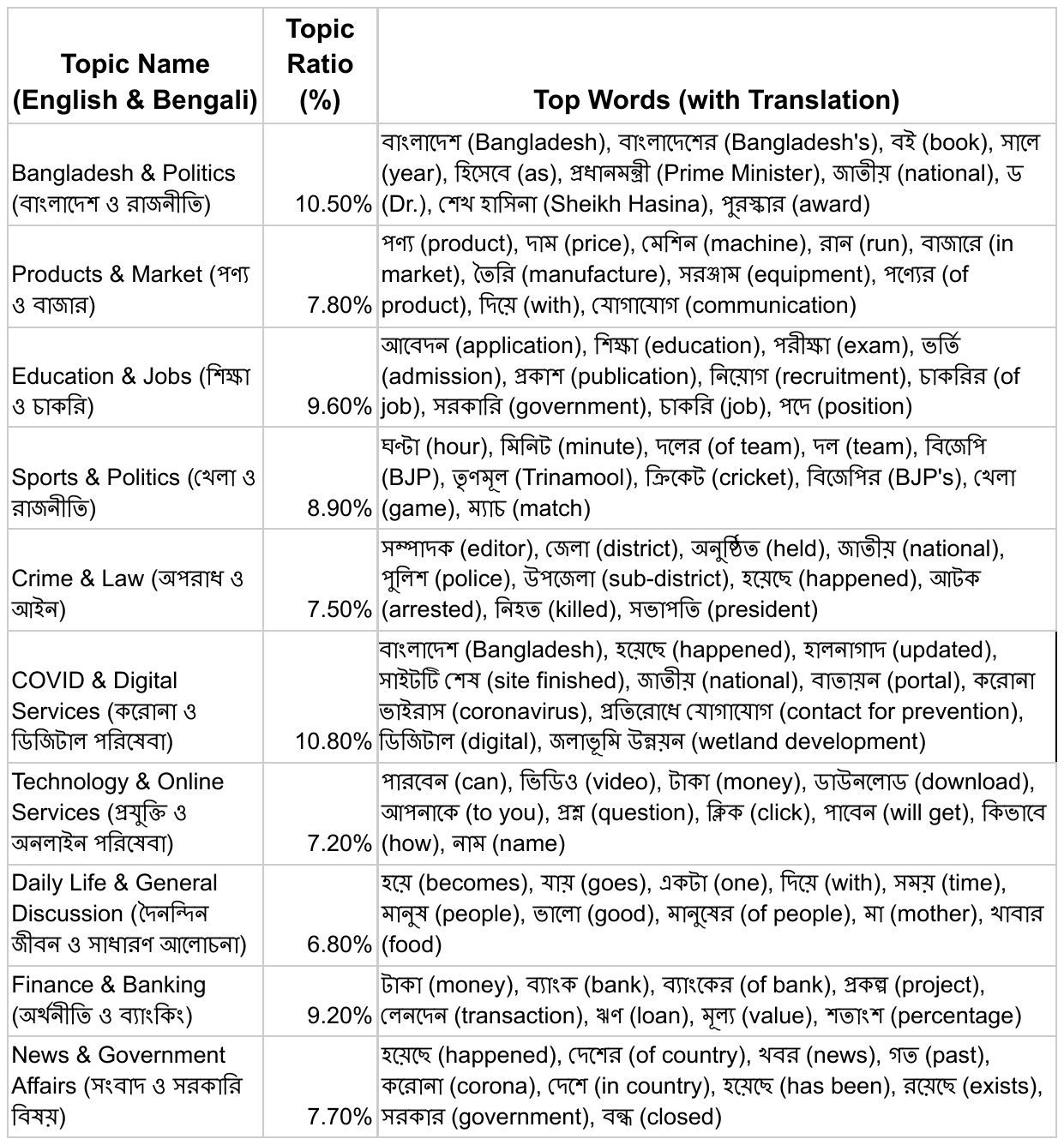}
    \caption{Topic modelling results for Bengali language in Chitrakshara-IL dataset.}
    \label{fig:topics-ban}
\end{figure*}

\begin{figure*}[htbp]  
    \centering
    \includegraphics[page=3, width=0.8\linewidth]{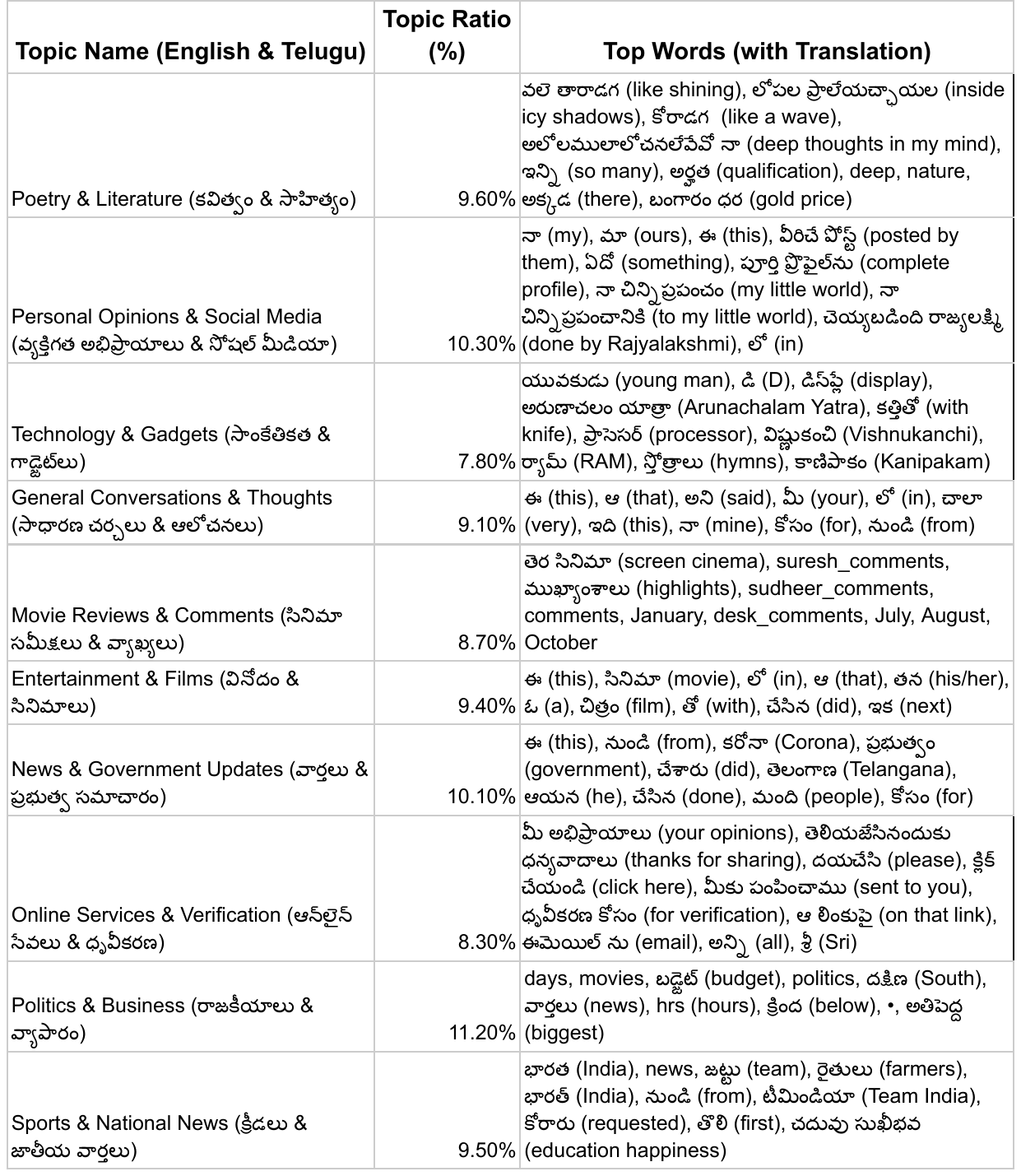}
    \caption{Topic modelling results for Kannada language in Chitrakshara-IL dataset.}
    \label{fig:topics-kan}
\end{figure*}

\begin{figure*}[htbp]  
    \centering
    \includegraphics[page=4, width=0.8\linewidth]{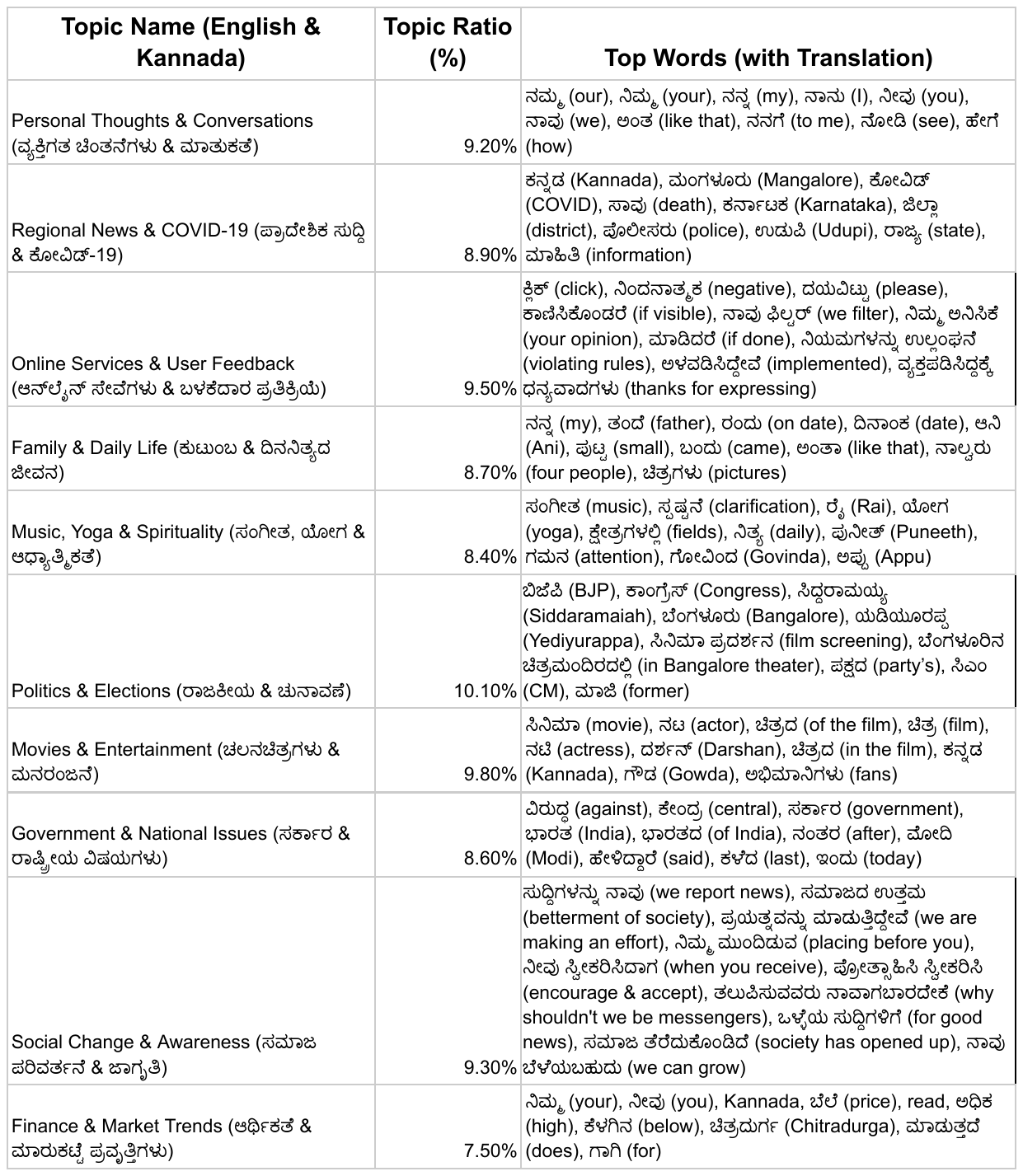}
    \caption{Topic modelling results for Telugu language in Chitrakshara-IL dataset.}
    \label{fig:topics-tel}
\end{figure*}

To gain a deeper understanding of the dataset's thematic structure, we apply Latent Dirichlet Allocation (LDA) \cite{blei2003latent}, a widely used probabilistic topic modeling technique. LDA helps uncover latent topics by analyzing word distributions and estimating their proportions across the dataset. Figures \ref{fig:topics-hindi}, \ref{fig:topics-ban}, \ref{fig:topics-kan} and \ref{fig:topics-tel} present topic modeling results for Hindi, Bengali, Telugu and Kannada  datasets, respectively. Each table provides both a broad categorization and a fine-grained breakdown of topics, facilitating a comparative analysis across languages. Our findings indicate a rich diversity of themes, including Politics, Entertainment, Health, Religion, and Technology, with certain domain-specific trends. Notably, journalism-related content appears frequently, suggesting that news articles constitute a significant portion. This pattern is consistent with trends observed in large-scale textual corpora, where online news sources contribute extensively to publicly available data.



    
\begin{table}[h]
    \centering
    \begin{tabular}{rll}
        \toprule
        Rank & Domain Name & \# Docs \\
        \midrule
        1 & hi.medindia.net & 346,395 \\
        2 & hindi.news18.com & 320,997 \\
        3 & zeenews.india.com & 307,467 \\
        4 & hindi.alibaba.com & 278,389 \\
        5 & amarujala.com & 273,655 \\
        6 & thamilislam.blogspot.com & 260,763 \\
        7 & jmi.ac.in & 249,841 \\
        8 & dbhwd.portal.gov.bd & 234,413 \\
        9 & esakal.com & 226,417 \\
        10 & cpiup.blogspot.com & 218,939 \\
        11 & patrika.com & 216,397 \\
        12 & hindi.oneindia.com & 213,376 \\
        13 & newsnationtv.com & 197,463 \\
        14 & hindi.webdunia.com & 177,184 \\
        15 & hindutamil.in & 175,280 \\
        16 & vikatan.com & 174,547 \\
        17 & jagran.com & 166,565 \\
        18 & maalaimalar.com & 162,471 \\
        19 & khabar.ndtv.com & 158,986 \\
        20 & newstracklive.com & 153,967 \\
        21 & aajtak.intoday.in & 147,650 \\
        22 & myupchar.com & 146,792 \\
        23 & bhaskar.com & 145,080 \\
        24 & aajtak.in & 132,935 \\
        25 & dailythanthi.com & 130,525 \\
        26 & tamil.oneindia.com & 127,532 \\
        27 & livehindustan.com & 127,211 \\
        28 & raji-rajiworld.blogspot.com & 125,848 \\
        29 & malayalam.oneindia.com & 119,674 \\
        30 & kannada.oneindia.com & 114,459 \\
        31 & gujarati.oneindia.com & 108,781 \\
        32 & navbharattimes.indiatimes.com & 108,541 \\
        33 & pustak.org & 106,866 \\
        34 & telugu.oneindia.com & 106,284 \\
        35 & bengali.oneindia.com & 104,611 \\
        36 & anandabazar.com & 103,539 \\
        37 & lokmat.news18.com & 103,088 \\
        38 & udayavani.com & 97,609 \\
        39 & origin-www.amarujala.com & 95,979 \\
        40 & abpnews.abplive.in & 94,350 \\
        41 & tv9marathi.com & 90,914 \\
        42 & celebrity.astrosage.com & 88,193 \\
        43 & ndtv.in & 86,399 \\
        44 & loksatta.com & 85,491 \\
        45 & newstrack.com & 84,154 \\
        46 & vivalanka.com & 84,075 \\
        47 & prabhasakshi.com & 83,619 \\
        48 & hindi.newsbytesapp.com & 82,588 \\
        49 & mathrubhumi.com & 79,742 \\
        50 & m.jagran.com & 78,261 \\
        \bottomrule
    \end{tabular}
    \caption{Top 1-50 URL domain names by number of documents in Chitrakshara-IL dataset.}
    \label{tab:top_domains}
\end{table}

\begin{table}[h]
    \centering
    \begin{tabular}{rll}
        \toprule
        Rank & Domain Name & \# Docs \\
        \midrule
        51 & bhopalsamachar.com & 77,566 \\
        52 & tamil.news18.com & 76,878 \\
        53 & india.com & 72,341 \\
        54 & origin1qaz2wsx-hindi.webdunia.com & 71,966 \\
        55 & cgkhabar.com & 69,293 \\
        56 & mpbreakingnews.in & 68,935 \\
        57 & ap7am.com & 68,798 \\
        58 & lion-muthucomics.blogspot.com & 68,121 \\
        59 & mumbailive.com & 67,404 \\
        60 & hi.topwar.ru & 64,629 \\
        61 & newstm.in & 62,938 \\
        62 & hi.forvo.com & 62,270 \\
        63 & thejasnews.com & 62,063 \\
        64 & asianetnews.com & 61,024 \\
        65 & globaltamilnews.net & 60,971 \\
        66 & khaskhabar.com & 60,394 \\
        67 & naturalfoodworld.wordpress.com & 57,818 \\
        68 & hmtvlive.com & 57,770 \\
        69 & hindi.latestly.com & 57,677 \\
        70 & specialcoveragenews.in & 56,047 \\
        71 & ujjawalprabhat.com & 55,647 \\
        72 & pravakta.com & 55,429 \\
        73 & bn.fanpop.com & 55,284 \\
        74 & rokomari.com & 55,252 \\
        75 & matrubharti.com & 54,751 \\
        76 & tv9hindi.com & 54,009 \\
        77 & navodayatimes.in & 54,009 \\
        78 & pricedekho.com & 53,973 \\
        79 & sharechat.com & 53,959 \\
        80 & bharatkhabar.com & 53,918 \\
        81 & hindi.catchnews.com & 53,878 \\
        82 & ek-shaam-mere-naam.in & 53,295 \\
        83 & hindi.asianetnews.com & 52,033 \\
        84 & hindi.siasat.com & 51,838 \\
        85 & dinamalar.com & 51,564 \\
        86 & bsb.portal.gov.bd & 51,367 \\
        87 & merisaheli.com & 50,811 \\
        88 & varthabharati.in & 50,529 \\
        89 & upuklive.com & 50,410 \\
        90 & sarita.in & 49,967 \\
        91 & mymahanagar.com & 49,775 \\
        92 & swadeshnews.in & 49,474 \\
        93 & dw.com & 48,642 \\
        94 & thewirehindi.com & 48,265 \\
        95 & earchive.amarujala.com & 47,870 \\
        96 & marathi.webdunia.com & 47,628 \\
        97 & copypastelove.org & 46,892 \\
        98 & bansalnews.com & 46,508 \\
        99 & maayboli.com & 45,694 \\
        100 & liveaaryaavart.com & 45,218 \\
        \bottomrule
    \end{tabular}
    \caption{Top 51-100 URL domain names by number of documents in Chitrakshara-IL dataset.}
    \label{tab:top_domains2}
\end{table}

\begin{figure*}[htbp]
    \centering
    \includegraphics[width=\linewidth]{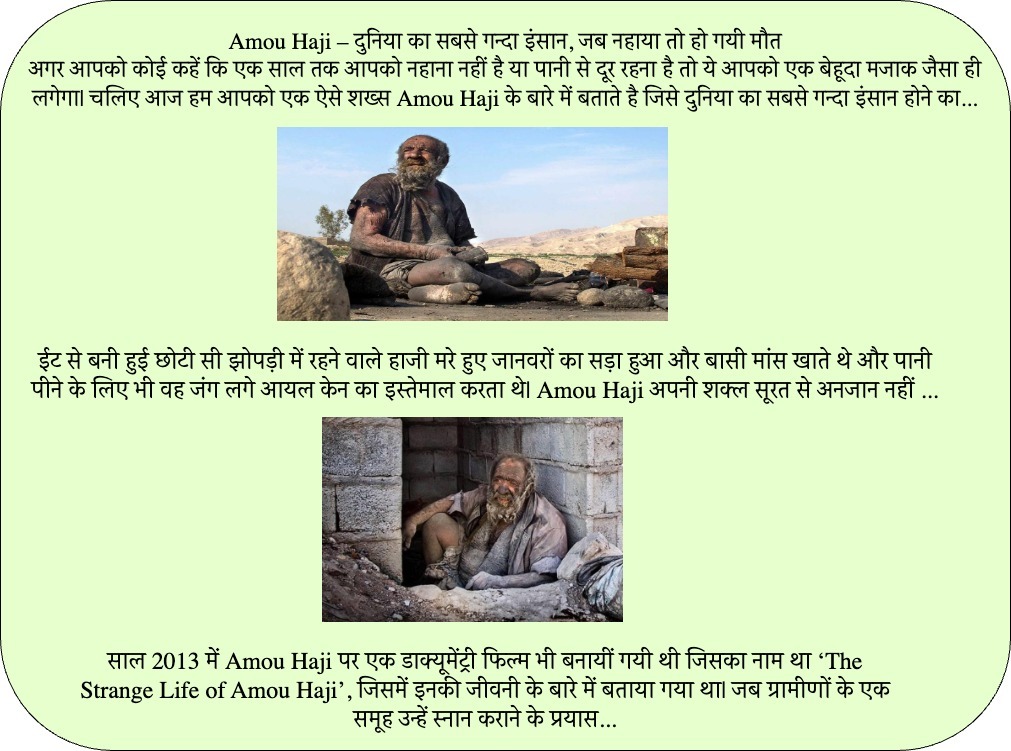}
    \caption{Example Interleaved document for Hindi}
    \label{fig:hindi-interleaved}
\end{figure*}

\begin{figure*}[htbp]    
    \centering
    \includegraphics[width=\linewidth]{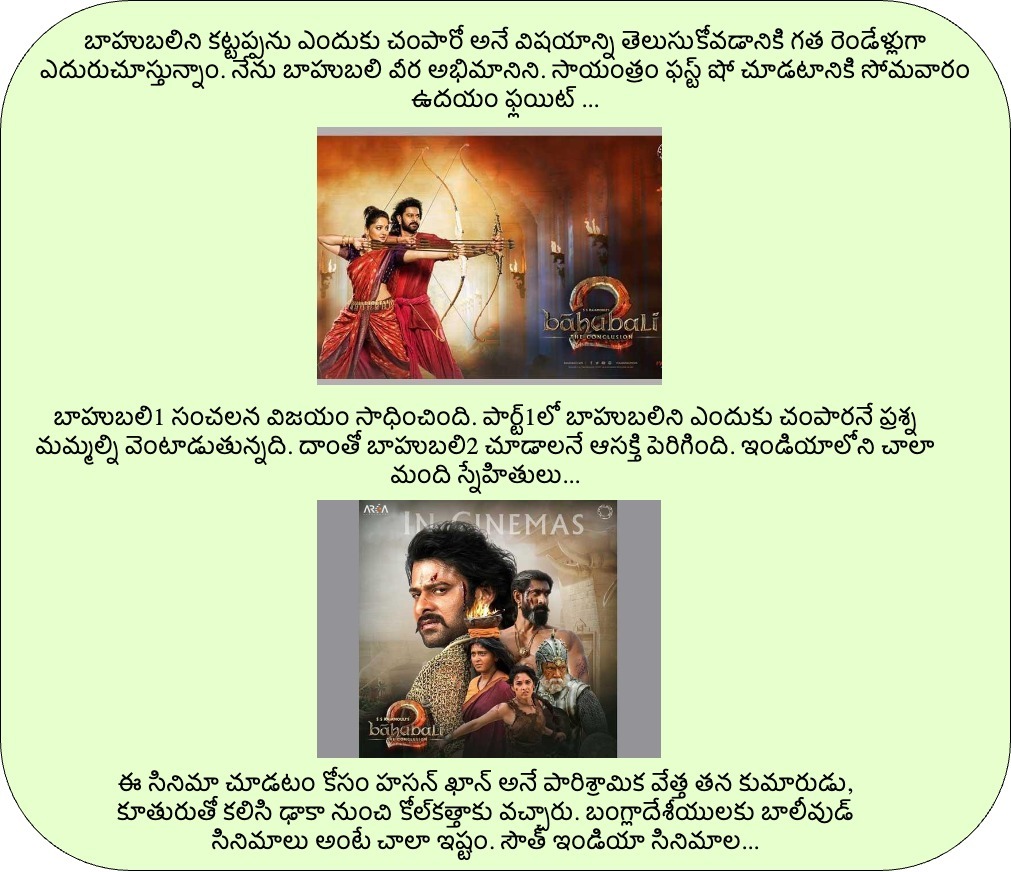}
    \caption{Example Interleaved document for Telugu}
    \label{fig:telugu-interleaved}
\end{figure*}

\begin{figure*}[htbp]    
    \centering
    \includegraphics[width=\linewidth]{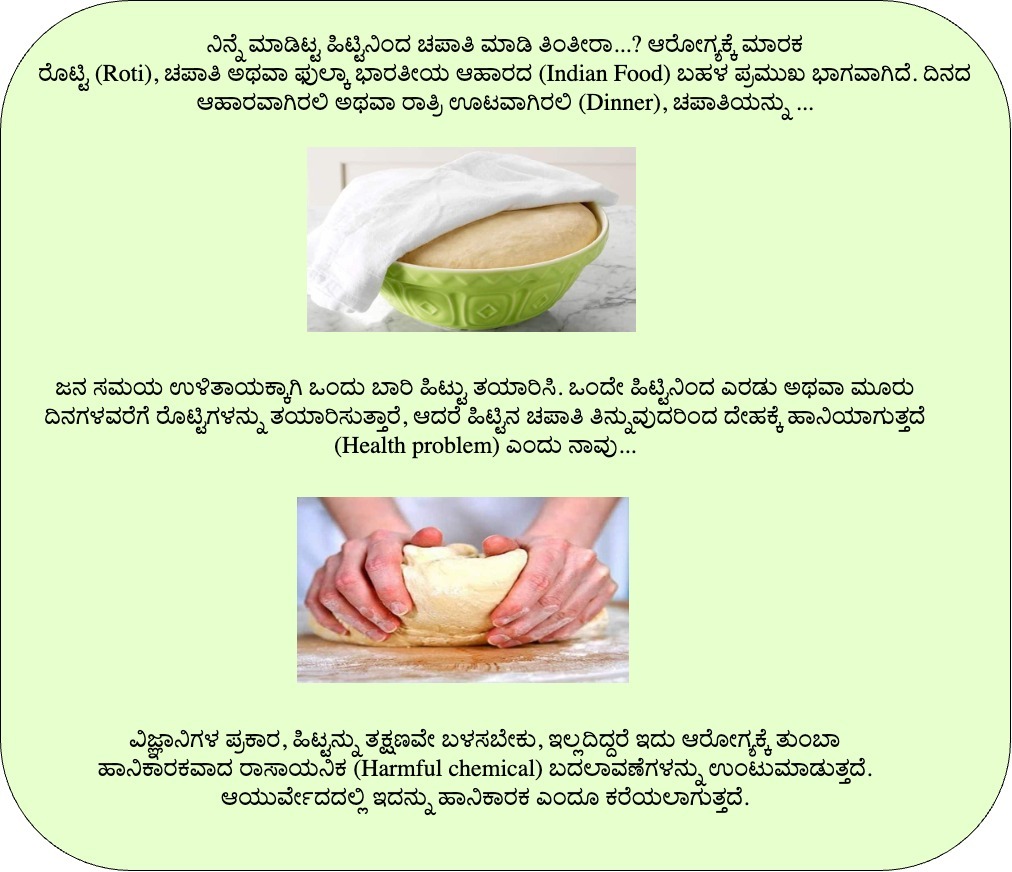}
    \caption{Example Interleaved document for Kannada}
    \label{fig:Kannada-interleaved}
\end{figure*}

\begin{figure*}[htbp]    
    \centering
    \includegraphics[width=\linewidth]{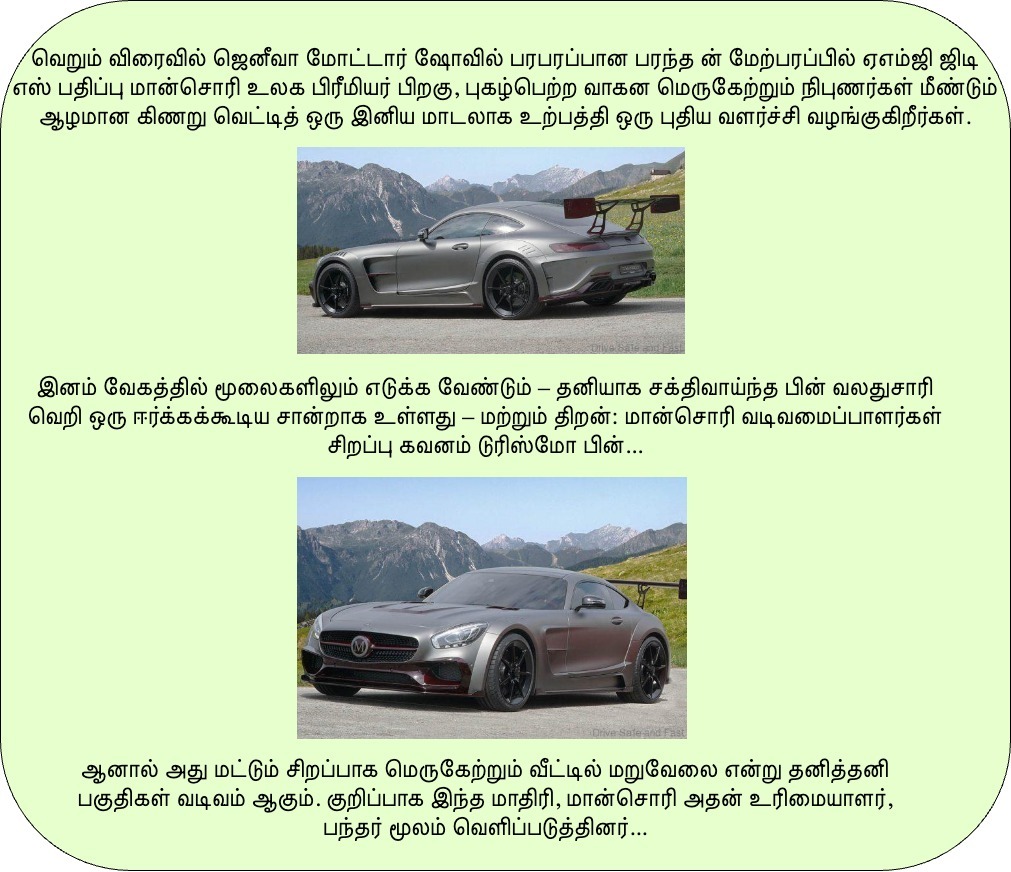}
    \caption{Example Interleaved document for Tamil}
    \label{fig:tamil-interleaved}
\end{figure*}

\begin{figure*}[htbp]    
    \centering
    \includegraphics[width=\linewidth]{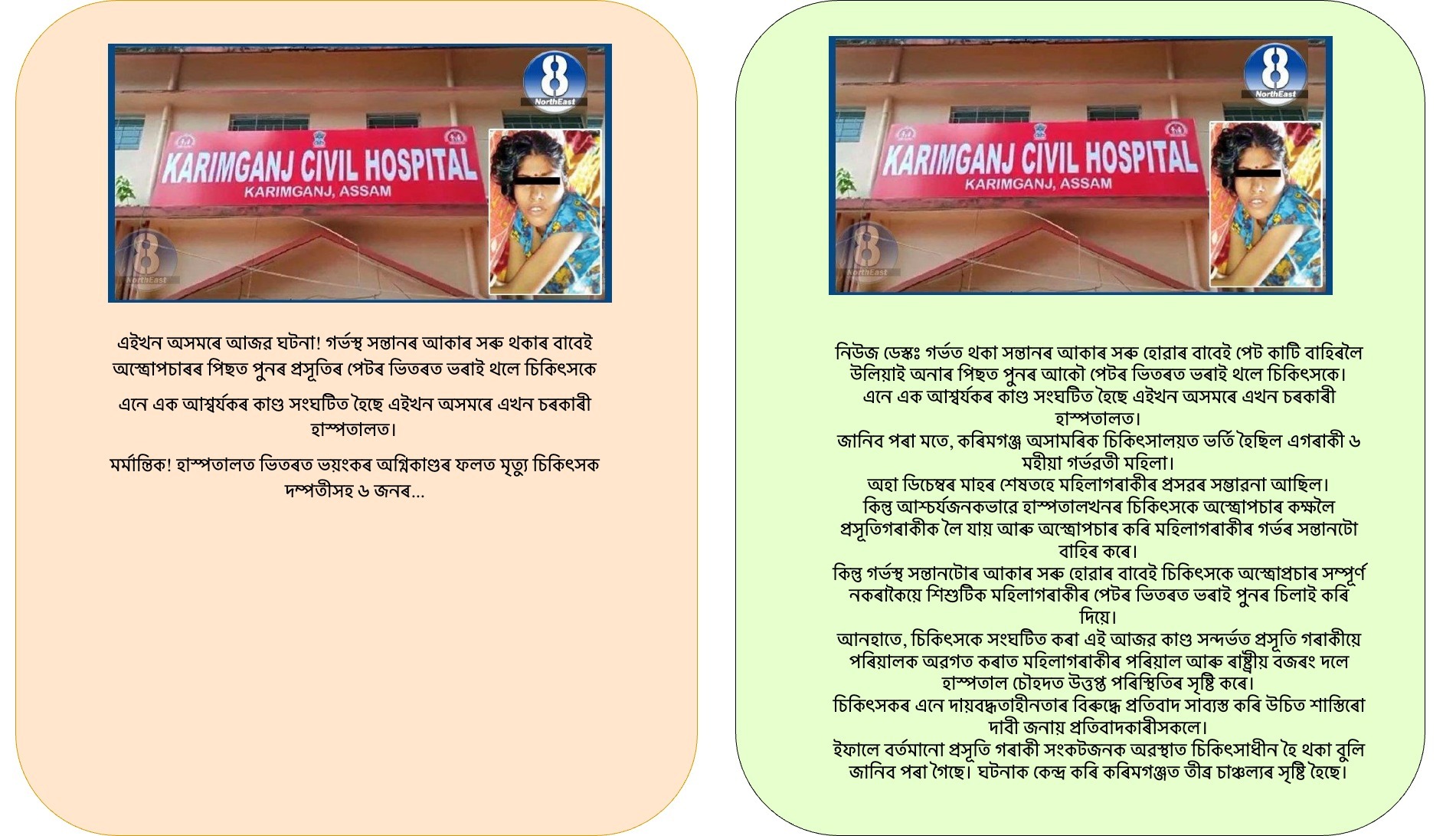}
    \caption{Comparison of the same interleaved document retrieved from mOSCAR against Chitrakshar-IL pipeline}
    \label{fig:comparison-interleaved}
\end{figure*}

\end{document}